# Atherosclerotic carotid plaques on panoramic imaging: an automatic detection using deep learning with small dataset

Lazar Kats*, Marilena Vered*, **, Ayelet Zlotogorski-Hurvitz*, ***, Itai Harpaz****

*Department of Oral Pathology, Oral Medicine and Maxillofacial Imaging, School of Dental Medicine, Tel Aviv University, Tel Aviv, Israel
**Institute of Pathology, The Chaim Sheba Medical Center, Tel Hashomer, Ramat Gan, Israel
***Department of Oral and Maxillofacial Surgery, Rabin Medical Center, Beilinson Campus, Petah Tikva, Israel
****Private Computer Practice, Tel Aviv, Israel

*Corresponding author: Lazar Kats, Department of Oral Pathology, Oral Medicine and Maxillofacial Imaging, School of Dental Medicine, Tel Aviv University, Tel Aviv 69978, Israel, Tel: +972-3-6409305, Fax: +972-3-6409250, Email: lazarkat@post.tau.ac.il

Keywords: Stroke, deep learning, panoramic imaging, neural network, atheroma, atherosclerotic carotid plaques

## Abstract

Stroke is the second most frequent cause of death worldwide with a considerable economic burden on the health systems. In about 15% of strokes, atherosclerotic carotid plaques (ACPs) constitute the main etiological factor. Early detection of ACPs may have a key-role for preventing strokes by managing the patient a-priory to the occurrence of the damage. ACPs can be detected on panoramic images. As these are one of the most common images performed for routine dental practice, they can be used as a source of available data for computerized methods of automatic detection in order to significantly increase timely diagnosis of ACPs. Recently, there has been a definite breakthrough in the field of analysis of medical images due to the use of deep learning based on neural networks. These methods, however have been barely used in dentistry. In this study we used the Faster Region-based Convolutional Network (Faster R-CNN) for deep learning. We aimed to assess the operation of the algorithm on a small database of 65 panoramic images. Due to a small amount of available training data, we had to use data augmentation by changing the brightness and randomly flipping and rotating cropped regions of interest in multiple angles. Receiver Operating Characteristic (ROC) analysis was performed to calculate the accuracy of detection. ACP was detected with a sensitivity of 75%, specificity of 80% and an accuracy of 83%. The ROC analysis showed a significant Area Under Curve (AUC) difference from 0.5. Our novelty lies in that we have showed the efficiency of the Faster R-CNN algorithm in detecting ACPs on routine panoramic images based on a small database. There is a need to further improve the application of the algorithm to the level of introducing this methodology in routine dental practice in order to enable us to prevent stroke events.

**Introduction**

Stroke is the second most frequent cause of death worldwide.[1,2] Patients who suffered strokes must undergo hospitalization followed by a period of rehabilitation, which often ends with partial recovery and/or permanent disabilities. The economic burden of strokes and their sequela on the health systems is considerable.[3]

Atherosclerotic carotid plaques (ACPs) constitute one of the main etiological factors of stroke, in about 15% of the cases.[4] There is a direct relationship between the degree of stenosis of the extracranial carotid artery, which is caused by ACP, and the risk of strokes.[5] Studies have shown that the annual risk of stroke is less than 1% among neurologically asymptomatic patients with less than 75% stenosis, however, the risk greatly increases to 2–5% in the group of patients with stenosis of more than 75%.[3,6]

Early detection of ACP is a key factor for preventing strokes by managing the patients a-priory to the occurrence of the damage.[7] ACP is relatively well diagnosed by various methods of medical imaging, such as Doppler ultrasound, computed tomography angiography (CTA) and magnetic resonance angiography (MRA).[3]

One of the most basic images, on which a pathology suspected for ACP can be detected, is the panoramic image. This is one of the most common images in dental practice. In addition to jaws and teeth, the area of the panoramic image includes also other anatomical zones, such as the area of the spine and adjacent tissues. ACP is formed initially in the field of the bifurcation of the carotid artery and when calcified enough, it becomes visible on the panoramic image.[8,9] According to studies, more than 3% of the ACPs can be detected on panoramic images of individuals over 55 years-old without a history of stroke or transient ischemic attack (TIA), and in more than 22% of the panoramic images among individuals who already have suffered from stroke or TIA.[10,11] In addition, a strong association between an increased frequency of ACP detection and age, was found.[10]

On regular panoramic x-rays, the field of bifurcation of the carotid artery is usually displayed in the area of the soft tissues of the neck, adjacent to or between the C3 and C4 cervical vertebrae. This area is clearly visible on the images, but, unfortunately, it is prone to be under-examined by dental practitioners. Whenever ACPs are identified on the panoramic images, the patient must be immediately referred for additional non-invasive examinations, such as Doppler ultrasound, and depending on the degree of stenosis of the carotid artery, conservative or surgical treatment is recommended, thus significantly reducing the risk of stroke.[12,13]

There are many studies demonstrating the effectiveness of computer-aided diagnostics in the analysis of medical images. Recently, there has been a definite breakthrough in the field of analysis of medical images due to the use of deep learning based on neural networks. In line with this, advances, at least at research level, have been made in the analysis of images of the lungs,[14] brain and breast.[15,16] Additionally, various techniques of machine learning, including neural networks, are being used for the detection of ACPs.[17-20] Specifically, in dentistry, there are two related studies, in which an automatic detection method using the top-hat filter was used for identifying carotid artery calcifications on dental panoramic x-rays.[21,22] In another study, a method based on the grayscale gradient has been suggested.[23] It should be emphasized that most studies in this field demand the use of large numbers of images. Acquiring large numbers of medical and especially dental images, is a most challenging task. In order to overcome this obstacle, we presently used a small number of dental images with accurate pre-processing segmentation. We aim to present a pioneering study that used neural networks for the automatic detection of ACPs based on a small dataset of panoramic images.

**Methods**

Network Architecture

For deep learning, we used the Faster Region-based Convolutional Network (Faster R-CNN) proposed by S. Ren and al.,[24] which is a combination between the Region Proposal Network (RPN) and the Fast R-CNN model. The base CNN used in our model was the Resnet-101. Further details of the structure of the network can be found in the original article and source code.[24,25] The simplified model structure is illustrated in Fig. 1.

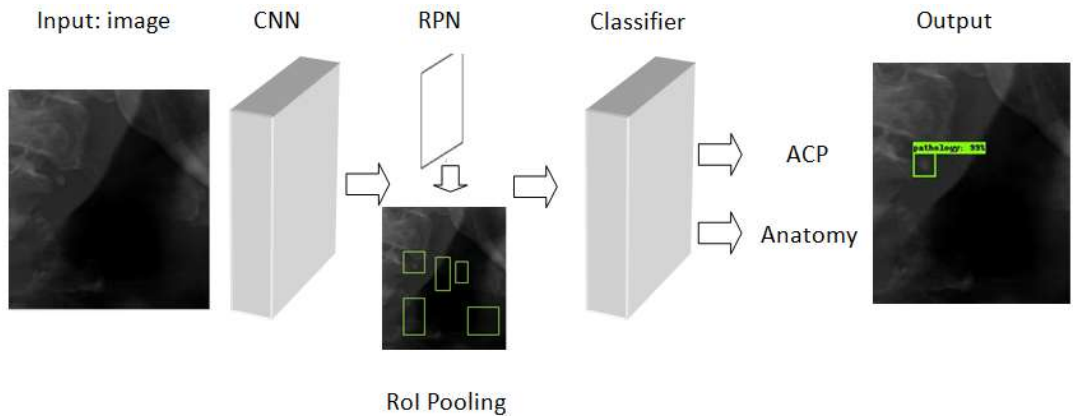

Fig. 1: Simplified model structure of the Atherosclerotic Carotid Plaque (ACP) detection using Faster R-CNN

Dataset

We used in-house dataset created at the School of Dental Medicine, Tel Aviv University. The Study was approved by the Ethical Committee of Tel Aviv University, May 10$^{th}$, 2018. The board waived the need for patient consent.

The dataset contained 65 Panoramic x-ray images from 2 different devices: CS 8100 Digital Panoramic System (Carestream Dental LLC, Atlanta, USA) and Planmeca ProMax 2D (Planmeca, Helsinki, Finland). Marking of the ACP was performed in a coordinated manner by two specialists in oral medicine and maxillofacial radiology. The dataset was challenging due to the large variability in ACP size, shape, intensity as well as varying ranges of contrast and sharpness of the images produced by the different devices.

Data augmentation and training

From each image, we automatically extracted left and right regions of interest (ROIs), where boundaries of ROI were the same for all images. These were calculated from the training of the data based on the location of ACP. As there was a small amount of available training data, we used data augmentation by changing the brightness and randomly flipping and rotating cropped ROIs to multiple angles (Fig. 2). This allowed the network to "learn" from the larger variability of examples and prevented overfitting. The input images and their corresponding segmentation maps were used to train the network in a mini-batch manner. In this way, 70% of the data were randomly chosen for training, 10% for validation and accuracy was tested on 20%. Gradient descent computation and updates were carried out by stochastic gradient descent optimizer (SGD). The model was trained using NVIDIA GeForce GTX 1080 GPU.

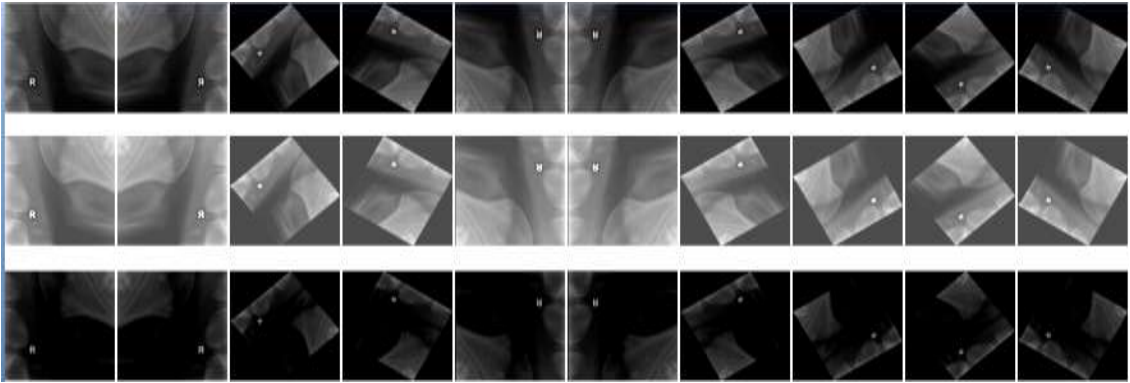

Fig. 2: Examples of data augmentation showing the effect of changing the brightness and randomly flipping and rotating cropped regions of interest

Statistical analysis

Receiver Operating Characteristic (ROC) analysis was performed to calculate the detection accuracy (null hypothesis: true area= 0.5), confidence interval (CI) 95%. The SPSS software, version 17.0 (Chicago, IL, USA) was used for the statistical analysis.

**Results**

We presently used images with the pathology (67%) and images with normal anatomy. An example of ACP detection process is illustrated in Fig.3.

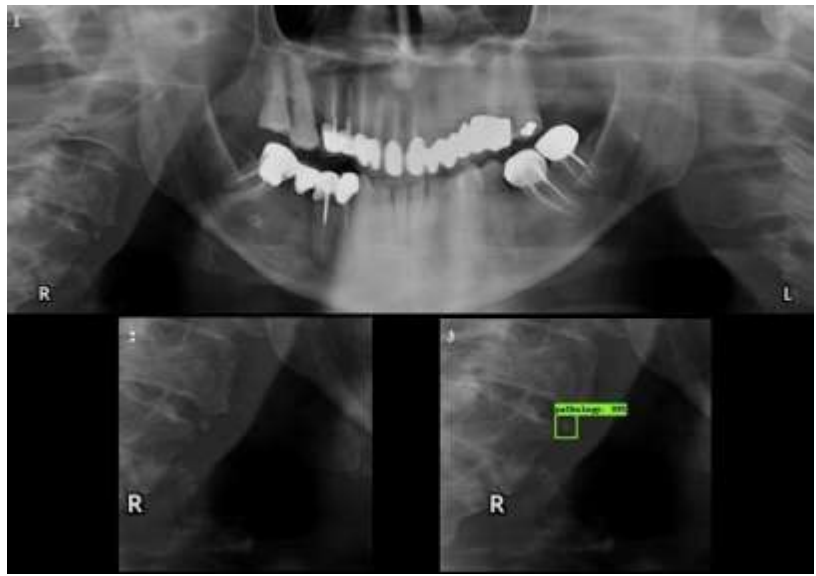

Fig. 3**:** Example of detection of an Atherosclerotic Carotid Plaque (ACP). 1. original panoramic image; 2. cropped image with a region of interest; 3. the outcome of the machine processing with detection and marking of the ACP

The sensitivity for the detection of ACP was 75% and the specificity was 80%, with an accuracy of 83%. The ROC analysis showed a significant Area Under Curve (AUC) different from 0.5 (null hypothesis) (Fig. 4).

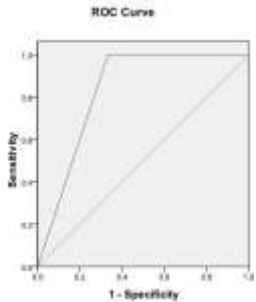

Fig. 4: ROC curve, AUC= 0.83

**Discussion**

In this study, we used the Faster R-CNN neural network, which is considered the state of the art for the detection and classification of images.[26] To these days, there has been no research in the field of dentistry using this technique, although in general medicine there are a number of successful developments.[26,27] We presently showed that the use of the network yielded a relatively high level of detection of ACPs and "no detection" in cases of normal anatomy, achieving a sensitivity of 75% and specificity of 80%. The results of the ROC analysis showed an AUC of 0.83, which further confirms the quality of the binary classification and supports the appropriateness of this method. Panoramic images from two different devices without preliminary processing were presently used. The purpose of this approach was to approximate the experimental conditions of the algorithm processing to those occurring in routine clinical settings.

Neural networks undergo a "learning phase", which requires a large number of images. In the field of automatic image analysis there is a general difficulty in collecting the database. One of our central tasks was to test the possibility of preparing a neural network for the subsequent detection process based on a small dataset containing only 65 panoramic images. Therefore, in order to substantially increase the dataset, various augmentations were performed, including rotations, flipping, and brightness variations, as well as various combinations of these procedures. This allowed us to increase by hundreds of times the existing number of samples for training.

7The radiographic image of ACPs demonstrates a great variety in terms of boundary contours, which are usually irregular. In addition, the lesions can consist of many small components. These features might severely complicate the segmentation, which needs a clear delineation of the boundaries. Under these circumstances, marking ACPs on panoramic images using a modification of a bounding box instead of the lesions contour, seems to be advantageous. Despite the absence of clearly delineated boundaries, the bounding box area containing the ACPs can be easily visualized.

Another aspect to be thoroughly considered when using automatic image analysis, and which inevitably adds some complexity to the system, is the differential diagnosis of possibly other calcified anatomical structures in the same region where ACPs occur. Basically, this includes the calcified triticeous cartilage and less likely the superior horn of calcified thyroid cartilage.[28] In case of incorrect preparation of the dataset, errors in detection might result. Therefore, the preliminary designation of the ACP is of utmost importance. Due to these reasons, each image was double checked by two specialists in order to eliminate errors. However, it is important to mention that the panoramic image is a two-dimensional projection of a three-dimensional object that can only give a preliminary estimate and in case of a suspected pathology, an additional study is required, usually a three-dimensional imaging examination.

Using the proposed algorithm may be of central help in the daily work of the dental practitioners as it can significantly increase the chances of timely detection of ACPs and thus prevent possible strokes. It is important to emphasize that the panoramic x-ray image is routinely performed for the purpose of establishing dental treatment planning and once the pathologic finding of ACP exists, it can be readily detected on it. Unfortunately, as a result of the fact that ACPs are not in the specific area which dental practitioners usually focus on, the pathology remains undetected. However, if the detection capabilities can be improved to a sufficient diagnostic level by automatic detection analysis, we can propose a standardized flow-chart for the prevention of strokes (Fig. 5).

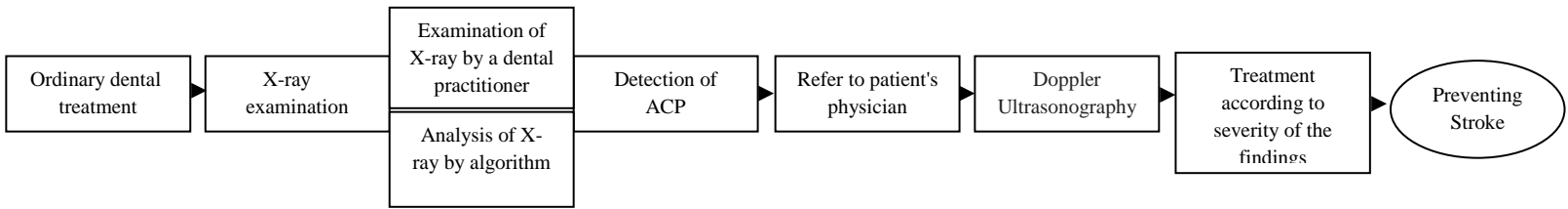

Fig. 5: A flow-chart of the algorithm for the prevention of stroke based on early detection of ACP in a panoramic image

**Conclusion**

Stroke is the second leading cause of death worldwide with ACP being one of the leading etiological causes. ACPs can be detected on panoramic images, one of the most common images in dentistry. Unfortunately, dental practitioners are usually unaware of the areas that are not directly related to their field of work. Having the ability to detect ACPs by means of an image analysis system may significantly improve the level of ACP diagnostics. Timely detection of ACP can prevent a subsequent stroke, which might have fatal outcomes or end in serious morbidity. Methods of machine learning, especially deep training, are being included in medical diagnostics at a higher pace nowadays. Our novelty lies in that we have showed the efficiency of the Faster R-CNN algorithm in detecting ACPs on routine panoramic images, which seems to be superior to the detection potential of the dental practitioners. Moreover, we were able to demonstrate that the use of a small dataset of images had the potential to yield satisfactory results. This can be further improved to the level of introducing this methodology in the routine dental practice.

**Disclosures**: none.